\documentclass{osa-article}

\journal{osac}

\articletype{Research Article}
\usepackage{multirow,array}
\usepackage{array,multirow,graphicx}
\usepackage{multirow}
\usepackage{comment}
\usepackage{ulem}
\usepackage{hyperref}
\usepackage{graphicx}
\usepackage{array}

\usepackage{soul,xcolor}
\newcommand\macolor{black}
\newcommand\hacolor{black}

\newcommand{\ma}{\textcolor{\macolor}}
\newcommand{\ha}{\textcolor{\hacolor}}

\newcolumntype{L}{>{\centering\arraybackslash}m{3cm}}
\begin{document}

\title{Deep intrinsic decomposition trained on surreal scenes yet with realistic light effects}

\author{Hassan A. Sial, Ramon Baldrich, and Maria Vanrell}

\address{Computer Vision Center, Edifici O, Campus UAB, Barcelona, Catalonia (Spain)}

\email{hasial@cvc.uab.cat} 

\homepage{http://cic.cvc.uab.es/hassan-ahmed/} 


\begin{abstract}
Estimation of intrinsic images still remains a challenging task due to weaknesses of ground-truth datasets, which either are too small, or present non-realistic issues. On the other hand, end-to-end deep learning architectures start to achieve interesting results that we believe could be improved if important physical hints were not ignored. In this work we present a twofold framework: (a) a flexible generation of images overcoming some classical dataset problems like larger size jointly with coherent lighting appearance; and (b) a flexible architecture tying physical properties through intrinsic losses. Our proposal is versatile, presents low computation time and achieves state-of-art results. 
\end{abstract}

\section{Introduction}

Despite its clear advantages for computer vision applications, intrinsic image decomposition in shading and reflectance components still remains a challenging task due to certain issues with datasets and a lack of physical considerations in the estimation methods. Before to revise different approaches, let us recall the intrinsic decomposition model initially proposed by Barrow and Tenenbaum in \cite{barrow1978recovering}, as the per pixel product of two components: 

\begin{equation} 
 I(x,y) = Ref(x,y) \cdot Sha(x,y)
\label{eq:intrinsicdecomposition}
\end{equation}

\noindent where $Ref$ and $Sha$ are the reflectance and shading component\ha{s}, respectively, and $(x,y)$ represents a pixel coordinates of the image, $I$. The problem pursuits to split a given single image in the estimation of shading and reflectance components, where shading is usually a grey level component capturing the effects caused by interactions between scene illumination, camera position and surface geometry, and reflectance is a color image capturing the chromatic properties of the surface object.

Unlike traditional prior work, current research in the estimation of these components is dominated by data driven methods. End-to-end deep learning architectures have shown promising results but they are still clearly improvable. In this paper we work on the hypothesis that ignoring model constrains is one of the reasons of still low quality results. Additionally, there is a common agreement in the fact that available ground-truth \ma{(GT)} image datasets are also responsible of poor performances. Realistic datasets either present few samples and lack of background interaction (MIT dataset)\cite{grosse09intrinsic} or they are large but not giving a full spatial coherence (IIW)\cite{bell14intrinsic}. On the other hand, synthetic datasets considerably increase the number of images but still other problems arise like low variability of backgrounds (Sintel)\cite{Butler:ECCV:2012:Sintel}, low interaction with these backgrounds like in Shapenet-based datasets \cite{Shi2017LearningNO,Baslamisli18}, \ma{and when this interaction increases, like in \cite{BaslamisliECCV2018,li2018cgintrinsics}, the physical coherence of lighting can be compromised.}

To deal with the above mentioned problems in this paper we propose a new synthetic image dataset of 25,000 images, at this moment. It is a Shapenet-based\cite{shapenet} dataset created in a sort of indoor-room ambient presenting a wide range of variations in shading, reflectance and cast shadows \ma{and with a physically consistent lighting}. Our ground-truth is based on the whole image and not just on some small masked object regions.  We also introduce a simple physical intrinsic loss on a versatile encoder-decoder architecture which reaches state of art results with much faster computation time compared to previous approaches, both traditional and deep learning methods. Our methodology to build the dataset can be easily extended for other related applications e.g light estimation shadow removal. The contributions of the proposed framework can be summarized in the next points:

\begin{itemize}
    \item A new synthetic dataset with two main advantages: (a) it has 25.000 images, which is much bigger that previous ones (Sintel 890, MIT 220, IIW 5000), making it more useful for training deep architectures; and (b) the proposed methodology provides a more realistic way to define light effects than the previous shapenet-based datasets.
    
    \item A new architecture for intrinsic image decomposition that reach state of art results for shading and reflectance. It is fast and easily extendible and imposes physical constraints on the loss function, which makes it to be very flexible to be extended on other constraints and being applied to other similar decompositions. 
    
    \item A whole methodology on building the dataset and the way the deep architecture is defined, that opens a vast set of opportunities to solve similar problems \ma{addressing the extraction of light components. This can be achieved} building realistic image configurations \ma{by easily tuning light effects accordingly with the task at hand, and generating their corresponding light components in the ground-truth. Having the ground-truth, it is easy to adapt the proposed architecture by introducing the corresponding losses fulfilling the decomposition model}. They can be applied to  applications \ma{focusing on estimation of light effects} like shadow detection and removal, estimation of color and position of light, etc. 
\end{itemize}

The paper is organized as follows. In section \ref{sec:previous} we do a literature review of most cited traditional approaches followed by more recent deep learning methods. Section \ref{sec:datasets} provides an overview of all available datasets in this domain and we conclude some convenient  properties to rationalise the construction of our dataset. Section \ref{sec:ourdataset} provides all the details on our dataset creation and its features. In Section \ref{sec:ourarchitecture} we explain our proposal for a deep learning architecture used in this paper. In section \ref{sec:experiments} we list all performed experiments, results and comparisons with previous results. Finally, section \ref{sec:conclusions} presents concluding remarks with some future work in this domain.

\section{Previous methods}\label{sec:previous}

Traditional work in this domain was focused on introducing physical priors on reflectance and shading and use optimization approaches to solve the decomposition problem. Earlier solutions in this domain were based on Retinex theory \cite{Land1971} that was based on the idea that shading corresponds to smooth variations while reflectance to abrupt changes in images. Thresholding image gradient can result in two estimations of reflectance and shading . This method was introduced for grayscale images and used later in different works \cite{horn1974determining, barrow1978recovering, Weiss2001, Tappen2005, Tappen2006}. Funt \textit{et al.}\cite{Funt1992} extended this idea for color images by assuming that shading is invariant to chromaticity.

 Posterior researchers added more physical constraints to this inverse problem, like introducing sparsity of reflectance \cite{Gehler2011, LShen2013}, or adding new priors on shape and lighting \cite{Barron2015}, or by adding higher-level color descriptors \cite{Serra2012}. In some works \cite{Barron2013,Lee2012,chen2013simple,jeonintrinsic2014} both RGB and depth information were used to predict intrinsic images, they provided promising results, but the lack of larger ground-truths stops the evaluation of these approaches. Others works focused on illumination varying image sequences from static video cameras \cite{Weiss2001,Matsushita2004,Laffont2015}, in these cases reflectance is constant through all the images and this multiple image-based approaches reduce complexity of intrinsic decomposition, these are very  specific application-dependent conditions that move away from the initial intrinsic problem from a single image. 
 
The broad success of Convolutional Neural Networks (CNN) in computer vision also tooks this community interest on solving this challenging task with supervised networks and by creating big and accurate amounts of data to train. The earlier approaches in this new framework started to appear in  2015. First, Narihira \textit{et al.} \cite{Narihira2015} which introduced \textit{Direct Intrinsics}, a two level end-to-end regression network to get shading and reflectance from single image. Their two stage network was inspired by Eigen \textit{et al.} \cite{Eigen2014DepthMP} that predicted depth from single image. At first level, the network tries to learn more global perspective of the image and at second level it tries to get more fine details from both input and first stage output. Second, Zhou \textit{et al.} \cite{Zhou2015LearningDR} presented a network to get relative reflectance between image pixel pairs trained on user annotated data between patches to have similar, larger or less relative reflectance built in \cite{bell14intrinsic}. The output of their network is a relative reflectance map for all image pairs, they used the same energy minimization function defined in \cite{bell14intrinsic} to get shading and reflectance from this map.  And third,  Shelhamer \textit{et al.} \cite{Shelhamer_2015_ICCV_Workshops} introduced a different methodology to use both depth and RGB to get intrinsic decompositon. They used fully convolutional network (FCN) \cite{Liu2015DeepCN} to predict depth from ground-truth image, and they used prior non deep learning methods designed in \cite{Barron2013,chen2013simple} to get shading and reflectance from both RGB input and predicted depth. In a similar approach, Kim \textit{et al.} \cite{Kim2016UnifiedDP} extended the idea to predict depth and intrinsic decomposition jointly from a deep architecture called joint conditional random field (JCNF) that shares convolutional activation's and layers between all $3$ tasks. JCNF learns to predict intrinsic images and depth jointly in a more correlated gradient domain. 

Afterwards, Nestmeyer and Gehler \cite{Nestmeyer2017ReflectanceAF} showed that prior knowledge on the problem domain can improve results for methods based on networks. They introduced a network which learns intrinsic decomposition based on the human judgements given in \cite{bell14intrinsic} and proposed a reflective filtering approach, that can be applied at the end of any network, improving prediction efficiency and computational cost.

\ha{Recently, some researchers tried to solve this problem by using more sophisticated deep learning architectures such as GAN by \cite{Goodfellow2014GenerativeAN}. Letrry \textit{et al.} in \cite{Lettry2018DARNAD} were the first to predict intrinsic images using a deep adversarial network.  At a first stage, CNN predicts only shading, and reflectance is considered to be element-wise division between the input and predicted shading. At a second stage, residual blocks are used to get final shading and reflectance from the first stage predictions. The novelty of this work is to introduce more physical constraint on loss term and merge data, gradient and adversarial losses together.}

\ha{Based on a different hypothesis that relies on the use of multiple images of the same scene, Ma \textit{et al.}\cite{Ma2018SingleII} used a Siamese network (introduced by \cite{Bromley1993SignatureVU}) to learn similar reflectance from multi illuminant images. The network is capable of learning to predict shading and reflectance from image pairs using  partial supervision i.e. without need of reflectance and shading ground-truth data. Weak supervision shows significant improvement in their framework. But, at learning time, the network needs image pairs under different lighting which are not always available. On the other hand, Li and Snavely in \cite{ZhenggiCVPR2018Learning} extended the previous work to train the network based on sequence of images from same scene with varying illumination. They collected a dataset of $145$ indoor and $50$ outdoor scenes and trained a network, that present some bias towards indoor scenes. It shows good performance, but with some blurring effects in the shading predictions, since the network learnt the invariance. All images in the sequence keep the reflectance constant, but a varying shading. This assumption was already posed in an early work on intrinsic decomposition by Weiss \cite{Weiss2001}.}

In spite of \ma{unsupervised approaches}, there is a common agreement in the fact that CNN-based approaches enhance the need for bigger image datasets. With this aim, Shi \textit{et al.} \cite{Shi2017LearningNO} introduced a large synthetic dataset and an encoder-decoder deep architecture to estimte shading, reflectance and specularity. This is the first deep learning architecture for non lambertian or non diffuse reflectance surfaces, where a specular component is added to the product model. This architecture presents one encoder and three decoders with shared connections. A similar approach was followed by \ha{Baslamisli} \textit{et al}. \cite{Baslamisli18} that introduced another synthetic dataset and two different deep learning architectures named as \textit{IntrinsicNet} and \textit{RetiNet} to get shading and reflectance from a single image. \textit{IntrinsicNet} is an encoder-decoder end-to-end network to predict shading and reflectance from a uniform architecture.  The \textit{RetiNet} has a more complicated scheme to incorporate traditional retinex theory \cite{Land1971} concepts in deep learning approaches. \textit{RetiNet} is a two-stage network, where the first stage learns shading and reflectance gradients, and the second stage combines both input image and first stage gradient information to get final shading and reflectance. Their claim is the use of more physical constrains on deep networks for a specific application. \ha{Similarly, Li and Noah \cite{li2018cgintrinsics} introduced a new synthetic dataset of $20,000$ images called CGintrinsics and use a U-Net based network with two decoders to predict shading and reflectance from single image. This network generalize well on complex scenes, but not providing good results on MIT dataset.}

More recently, Fan \textit{et al.} \cite{Fan2018RevisitingDI} introduced a multi-stage deep learning architecture. They use a pretrained Direct Intrinsics network \cite{Narihira2015} to get rough shading and reflectance at a first stage. In parallel they use another network to get a guidance map from the input image edges. In a later stage they use both the first stage Intrinsic decomposition and the guidance map to build a smooth final reflectance. Shading is an element-wise division between input image and final stage smooth reflectance. This network model is complex and is dataset-dependent.

All the above mentioned methods have been evaluated and some of them also trained on different available image datasets. In the next section we review existing datasets and we analyse and compare all of them, and we \ma{present our approach} to synthesize \ma{a} diverse datasets to \ma{tackle} the intrinsic decomposition \ma{of light components}.

\section{Datasets for Intrinsic decomposition}\label{sec:datasets}

The main problems regarding datasets for intrinsic image estimation swings between the realism of the physical lighting properties of the scene, and the number of images they provide to be enough to train deep architectures. The performance of trained architectures usually rely on a large number of training samples and a high variable appearance between them. 

\textit{MIT Intrinsic}\cite{grosse09intrinsic} was the first dataset in this domain, and it is a non synthetic dataset, it was captured under very controlled conditions. It has $20$ objects, captured in $11$ different lighting conditions, although only one shading (white object acquisition) and one reflectance pair is provided  the final dataset ground-truth. Different intrinsic image versions can be generated by introducing a scalar $\alpha$ that minimizes the difference, $I(x,y)-(\alpha Re(x,y)\cdot Sh(x,y))$, since even in such controlled environment the product model does not hold for all pixels in objects. The ground-truth is just provided for the pixels in the object mask.

\textit{MPI Sintel}\cite{Butler:ECCV:2012:Sintel} is a synthetic dataset based on an animated movie. It has $18$ scenes with $50$ frames each, except one that has $40$, totaling $890$ images. Sintel has been the first large dataset giving the opportunity to train deep architectures in the last years. However, it presents unnatural shading, and some color bias mostly on blue and brown colors. Thus, the generalization of networks trained on Sintel are affected by these color biases. And, like in all synthetic datasets, some erroneous pixels are found around boundary of objects.

\textit{IIW: Intrinsic Images in the Wild}\cite{bell14intrinsic} is a large and realistic dataset of 5230 images. It only provides sparse reflectance pairwise judgments as training data. These judgments does not present a spatial coherent map for reflectance. Consequently, networks trained using this dataset present too smooth reflectance estimation with lack of texture variation. 

\textit{MIII dataset} proposed by Beighpour \textit{et al.} \cite{Beigpour2016MultiviewMI} is the extension of the MIT Intinsic to a multi illuminant intrinsic dataset. This work is a mixture of a realistic and synthetic framework. First they do a $3D$ reconstruction of the objects and make a scene of multiple objects in a synthetic world. Objects are colored in a synthetic world and captured in different lighting conditions. Dataset has $5$ scenes and each scene is captured in $15$ different illumination conditions giving the total of $75$ images. Only one reflectance, specular and depth image is provided for each scene. Although their approach is very original and the dataset can be easily scaled by increasing light colors, objects and viewpoints, the $3D$ reconstruction based on computer vision algorithms of the objects is not perfect. Small differences between real and reconstructed object edges, can ultimately result in erroneous shading and reflectance in some parts of the image.

\textit{ShapeNet}\cite{Shi2017LearningNO} is the first large dataset in this field with $330,000$ images based on the Shapenet $3D$ objects dataset \cite{shapenet}. They used environmental maps in render software to create shading, reflectance and specular component of objects. The final ground-truth is only based on masked regions of objects without cast shadows information. 

\ma{\textit{Baslamisli \textit{et al.} in} \cite{Baslamisli18}, follow a similar approach to the previous one.} They created an intrinsic dataset of $20,000$ images called Shapenet-Intrinsic. Instead of using original shapenet object textures, they used homogeneous color reflectances for each object part to have more reflectance variation and to disassociate the shape from texture. The final dataset is again only based on masked object region  \ma{enlightened by environmental maps}.

\ha{\textit{Baslamisli \textit{et al.} in} \cite{BaslamisliECCV2018} present a new synthetic dataset of $35K$ images with new synthetic natural objects to form more complex natural scenes. The ground-truth not only provides intrinsic components but also annotations for semantic segmentation purpose. Images are created using sky HDR environmental maps with parallel lighting in the background . These maps introduce different  daylight conditions such as clear sky, cloudy, night etc. The whole foreground area has been considerably increased  with respect to the previous two mentioned datasets, sky portion areas are masked out in the ground-truth images.}   

\ha{\textit{CGIntrinsics}\cite{li2018cgintrinsics} is another recent synthetic ground-truth of $20,000$ images. They use the $3D$ indoor objects of SUNCG dataset \cite{song2016ssc} with 3D textures. It contains images representing complex indoor scenes of objects, combining a  mixture of indoor and outdoor illumination. Outdoor illumination sources are also based on HDR environmental maps. Both indoor and outdoor lighting sources are masked out in ground-truth.} 

\ha{\textit{InteriorNet} is a very large scale dataset recently presented in \cite{InteriorNet18}. It is formed by $20$ milions of images with an extense ground-truth for a variety of applications such as segmentation, object boundary detection, depth map estimation or motion blur removal. They also provide an interactive simulator ViSIM and the rendering software to randomly select objects or change lighting configuration beyond more tools. Although this dataset is not just made for intrinsic decomposition, it provides a good starting point to create a large photo-realistic and physically consistent dataset. In its current state the dataset does not provide intrinsic ground-truth data and the product model is not held.}  

Before to sum up on current datasets it is worth mentioning an additional way to solve the lack of data for training, it is using specific data-augmentation for intrinsic estimation, as suggested by Sial \textit{et al.}. in \cite{Sial2018}. They propose chromaticity rotation to add new reflectance images while keeping the same shading. It is a flexible augmentation strategy that can considerably increase the size of the dataset while reducing the problem of color bias in some datasets. 

To conclude the review of the different available datasets we provide a complete comparison in \ha{Table} \ref{tab:dataset_comparison} where we analyse the different datasets according to several properties we have organized in columns referring to: (a) the size or the number of available images with the corresponding ground-truth data; (b) if the dataset fulfills the product model given by \ha{Equation}\ref{eq:intrinsicdecomposition}; (c) if the full image can be used in the training, with no mask that reduces the number of samples and meaningful spatial coherence in the training stages; (d) if the ground-truth captures the influence of a diversified background that provoke a large diversity of lighting effects on the object surfaces; (e) if the ground-truth images present cast shadows that add realism to the full scene; \ma{(f) if the global lighting is physically consistent with real interaction between all the scene objects, environmental maps usually alter this coherence.}

\begin{table}[h!]
    \centering
    
\small{
\begin{tabular}{|l|c|c|c|c|c|c|} \hline
\multicolumn{1}{|c|}{Dataset} &

\begin{tabular}{@{}c@{\hspace{0.05in}}c@{}}\rotatebox[origin=c]{90}{Size} & \rotatebox[origin=c]{90}{$(\#Images)$}  \end{tabular} & 
\begin{tabular}{@{}c@{\hspace{0.05in}}c@{}} \rotatebox[origin=c]{90}{Model} & \rotatebox[origin=c]{90}{Fulfillment}  \end{tabular}  &
\begin{tabular}{@{}c@{\hspace{0.05in}}c@{}} \rotatebox[origin=c]{90}{Training on} & \rotatebox[origin=c]{90}{full Image}  \end{tabular}  &
\begin{tabular}{@{}c@{\hspace{0.05in}}c@{}} \rotatebox[origin=c]{90}{Diversified} & \rotatebox[origin=c]{90}{\hspace{0.1cm}Background\hspace{0.1cm}}  \end{tabular}  &

\begin{tabular}{@{}c@{\hspace{0.05in}}c@{}} \rotatebox[origin=c]{90}{Cast} & \rotatebox[origin=c]{90}{Shadows}  \end{tabular} &

 \begin{tabular}{@{}c@{\hspace{0.05in}}c@{}} \rotatebox[origin=c]{90}{Consistent} & \rotatebox[origin=c]{90}{Lighting}  \end{tabular}  \\   \hline
         MIT \textit{(Grosse et al. \cite{grosse09intrinsic})} & 220 & yes$^\star$ &  no & no & no & yes\\
         IIW \textit{(Bell et al. \cite{bell14intrinsic})} & 5230 & no & no & yes & yes & yes\\ \hline
         MIII \textit{(Beigpour et al. \cite{Beigpour2016MultiviewMI})} & 75 & yes & no & no & no & yes \\
         Sintel \textit{(Butler et al. \cite{Butler:ECCV:2012:Sintel})} & 890 & no & yes & yes$^\ddagger$ & yes & yes\\\hline
         ShapeNet \textit{(Shi et al. \cite{Shi2017LearningNO})} & 330K & yes &  no & yes & no & no\\
         Shapenet-Intrinsic (\textit{Baslamisli et al. \cite{Baslamisli18}}) & 20K & yes & no & yes &  no & no \\ 
         \textbf{Our dataset (SID)} & 25K & yes & yes & yes & yes & yes \\ \hline
        \ma{\textit{Baslamisli et al. \cite{BaslamisliECCV2018}}} & \ha{35K} & \ha{yes} & \ha{no$^\dagger$} & \ha{yes} &  \ha{yes} & \ha{no}\\
        \ma{CGintrinsics \ha{\textit{(Li and Snavely\cite{li2018cgintrinsics})} }} & \ha{20K} & \ha{yes} & \ha{no$^\dagger$} & \ha{yes} &  \ha{yes} & \ha{no}\\
        \ma{InteriorNet \ha{\textit{(Li et al.\cite{InteriorNet18}}) }} & \ha{20M} & \ha{no} & \ha{no$^\dagger$} & \ha{yes} &  \ha{yes} & \ha{no}\\
         \hline
    \end{tabular}}
    \caption{\ha{Comparison on current available dataset according to several properties. From left to write we account for: Number of images, GT perfectly fulfills the physical model, GT is on the full image or only a part, GT is presenting the influence of a diverse background, GT is presenting cast shadows apart from shading, and global image present physically consistent lighting. Meaning of special cases: ($\star$) MIT dataset generally fulfills product model by including a factor i.e. $I=\alpha(R\cdot S)$, but it does not completely hold for all images and have small deviation; ($\ddagger$) Sintel dataset present diverse backgrounds compared to the rest, but with a strong bias towards specific colors due to high correlation of a video sequences. ($\dagger$) Training area is large, but still does not cover the full image.}}
    \label{tab:dataset_comparison}
\end{table}

From \ha{Table}\ref{tab:dataset_comparison} we can conclude that the majority of available datasets are synthetic, only the first two rows correspond to realistic ones. MIT is the only realistic dataset that provides full reflectance and shading images, since IIW only give reflectance judgement data for specific pairs.  MII is a synthetic dataset carefully created for intrinsic decomposition, whereas Sintel was not created for this problem but has been one of the most used, since it was the first presenting an enough number of images to train deep architectures, its main problem is the high level of correlation between all the images. 

 \ma{The three subsequent rows in Table \ref{tab:dataset_comparison} correspond to Shapenet-based datasets, including our proposal. ShapeNet \cite{Shi2017LearningNO} emerged as a tool to create new larger datasets where synthetic objects are located in multiple different environmental maps. Following this idea, Baslamisli \textit{et al.} \cite{Baslamisli18}  used the same approach but using homogeneous reflectance for each object mesh. \ma{In both cases the ground-truth is just given by the object area. However, our proposal, also based on Shapenet, uses a single reflectance per mesh like in \cite{Baslamisli18}}, but substituting environmental maps by multiple elements in the scene surrounding the object that inserts a diversified background, extends the training area to the full image, and adds realism to light effects thanks to shadows and to the physical consistency on rendered light effects. This dataset is presented here as a tunable baseline to easily generate a high diversity of ligthing conditions, that can be adapted depending on the task at hand. We explain all the details on how this dataset is built in the next section.}

\ma{At the bottom of table \ref{tab:dataset_comparison} we have grouped 3 recent datasets that increase the complexity of the scenes, extend the ground-truth to larger areas that can contain cast shadows, but keeping environmental maps in some other parts, which can not be included in the ground-truth and provoke some lack of coherence in the global lighting of the scene. These datasets can be used for more generic applications where a high accuracy in the estimation of light conditions is not required.}

\section{Our dataset}\label{sec:ourdataset}

\ma{Here we propose a dataset that can be as large as needed, presents realistic light conditions and which can be easily adapted to different tasks addressed to the estimation of light components.} It is based on Shapenet objects \cite{shapenet} without textures like in \cite{Baslamisli18}, although they could be added. However the main difference is that we improve realism of light effects and training capabilities. We pursue a dataset with the following properties: (a) existence of cast shadows \ma{and global light consistency}; (b) keeping the influence of a varying surround; and (c) taking advantage of using the full image information for training.

To achieve the previous advantages we substitute environmental maps by multi-sided rooms with highly variable reflectances on the walls. \ha{Although the current dataset uses multi-sided flat walls, the shape of the rooms can be extended to multiple wall shapes, such as cylindrical or warped. In this work we focus on a simple version to start testing the approach on intrinsic decomposition.}  The main property of environmental maps is to get diverse lighting conditions, since they allow to assume a different light source at every point of the background, but the drawback of these maps is that they are unable to cast shadows \ma{and introduce some physical inconsistencies in the light interactions of the scene.}. A synthetic scene with walls covered by textured patterns and some point light sources at different random positions in the room can be a useful background to simulate a large number of scene images with the three proposed properties.

\begin{figure}[h]
\centering\includegraphics[width=13cm]{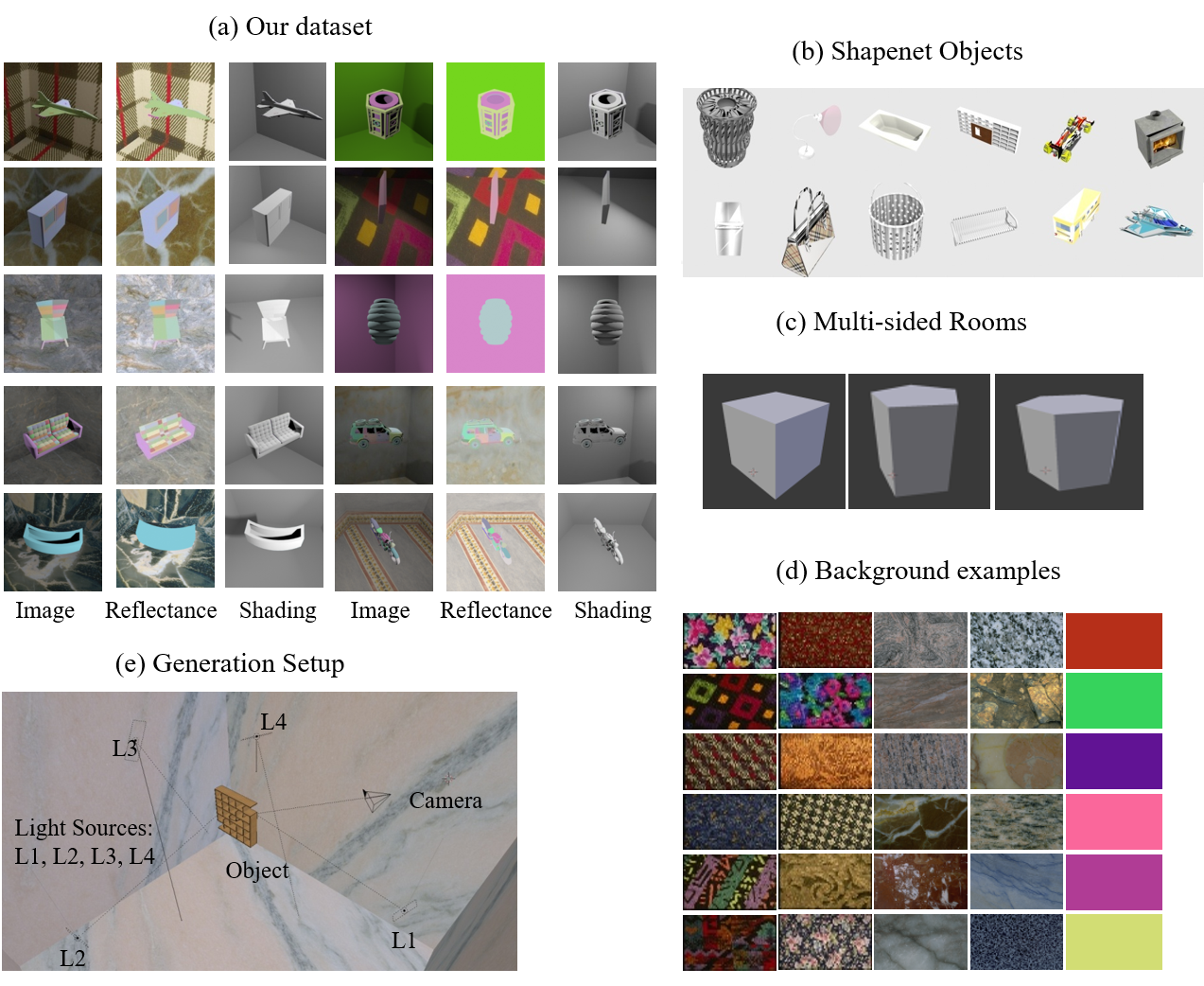}
\caption{Dataset Generation Setup}\label{fig:dataset}
\end{figure}

Our dataset has $25,000$ ground-truth images with intrinsic data for the whole image pixels. We used the open source Blender rendering engine to generate images, which can be used with multiple GPU's making rendering much faster. Below we list the key features of our dataset:
\\
\\
\textbf{Objects:} We used $12,500$ 3D objects from Shapenet\cite{shapenet}, randomly selected from several object categories such as bus, car, chair, sofa, airplane, pots, electronics etc. Similarly to \cite{Baslamisli18}, we also observed that object textures affected Blender shading, resulting in a wrong reflectance and shading decomposition in the generated ground-truth, to solve this issue, we also used a diffuse bidirectional scattering distribution function (BSDF) with random color and roughness values for each mesh texture in object. Roughness parameter controls how much light is reflected back from each object surface. Our dataset presents significant shading and reflectance variation across different objects and its surfaces. Objects are positioned in the image center. \\
\\
\textbf{Backgrounds:} To have diverse backgrounds of shading and reflectance, we defined $4$ to $6$ sided rooms (see \ha{Figure} \ref{fig:dataset}.(c)), where the number of flat walls was randomly selected for each image. These rooms provide a wide range of different geometric configurations with diverse light reflection conditions on the objects. Walls were colored with $50$ homogeneous color and $200$ variable set of textured patterns. Textured images were carefully selected from \ha{Corel dataset \cite{Corel}}, we choose the subset \textit{TexturePattern}, formed by fabrics, and \textit{Marbles}, with the aim to ensure they were flat surfaces with no shading effects projected on the image texture pattern (see \ha{Figure} \ref{fig:dataset}.(d)). In the current version of the dataset we have used a unique color or texture for all the walls in the scene, in order not to increase too much the variability of this first version, but this can be introduced as an additional parameter to increase the variability of the dataset.
\\
\\

\textbf{Generation setup:} \ma{To generate random scenes we locate the object at the center of the scene, where we also put the centroid of the volume formed by the walls, fixing the distance between the object and the walls within a range that ensures that all objects fit in the room. The camera is placed at a fixed distance from the center of the scene, the position is randomly selected on the defined semi-spherical surface. The camera image plane is orthogonal to the line from the center. Each object is captured twice, by using two camera positions separated by $180$ degrees in the horizontal pan axis of the semisphere, in this way, we can get two different views of the same object. To set the scene lighting we put $4$ white light sources with fixed location and orientation, we just randomly changed the intensity, that is what most affects the intrinsic shading. Two of the light sources having lower intensity range while higher for the two other to create more shading effects. We have used cycle rendering with GPU support to render images in blender.}

To sum up, the dataset proposed in this work is just a first version that can be easily extended by varying the multiple parameters we have introduced above, and which can be selected accordingly with the task we need to solve. The final scenes we create are somehow surreal, since images are presenting \textit{"airplanes flying in closed rooms decorated with old-fashioned wall-papers"}, but providing a wide range of realistic light effects with multiple color and shade interactions, thus we call it SID for \textit{Surreal Intrinsic Dataset}.

\ma{The proposed dataset is specially designed for intrinsic decomposition. The way in which it is designed makes it to be easily tunable, since the number of parameters is not very high and the computation time for generation is very low. Therefore, new datasets can be derived for other applications like: (a) \textit{Color constancy}, we just need to set the number of light sources we want and change the color, which requires to be stored in the ground-truth; (b) \textit{Shadow removal}, we need to separately synthesize shading with and without shadows and make the network to learn the difference between this two images; and (c) \textit{Specularity detection}, in this case we need to introduce non diffuse objects in the scene and generate their specular components for the ground-truth. Additionaly, complexity could be increased by adding more than a single object, or inserting 3D textures on objects and walls, the scenes can be modeled depending on the light effect to be estimated.}

\section{Our deep architecture}\label{sec:ourarchitecture}

We propose a deep architecture based on the following three criteria: (a) the use of a U-NET-based, Encoder-Decoder, architecture, which has been the most usual \cite{Shi2017LearningNO,Baslamisli18} and natural way to solve this pixel-wise regression problem; (b) increasing efficiency and speed properties by introducing split inception modules decreasing the number of parameters \cite{szegedy2016rethinking}; and (c) introducing physical constraints of the intrinsic decomposition model at the loss function like in  \cite{Lettry2018DARNAD,Baslamisli18}. We will refer to our architecture as IUI as for \textit{Inception U-Net} based for \textit{Intrinsic} image estimation.

\begin{figure}[h]
\centering\includegraphics[width=13cm]{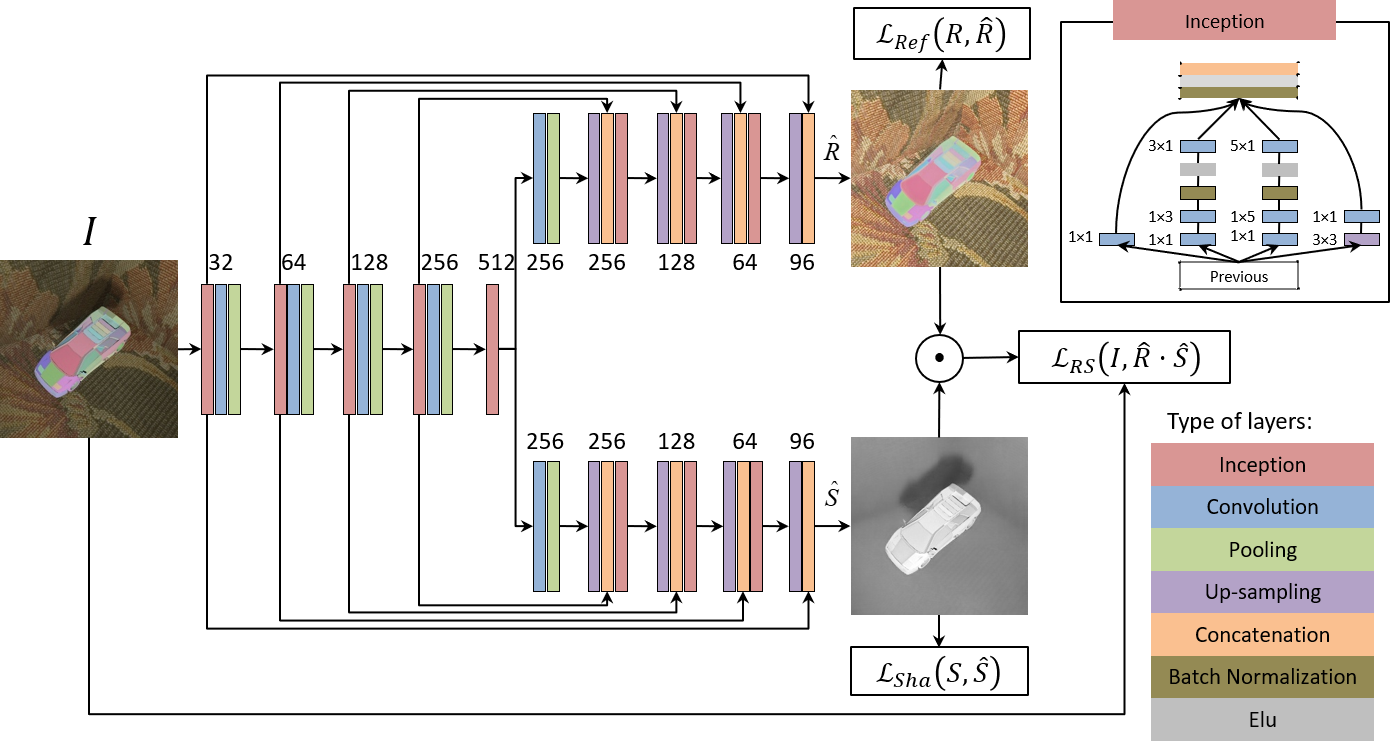}
\caption{IUI-Network architecture. One encoder and two decoders for reflectance and shading estimation. Three inter-related loss functions. Type of layers are indicated by a color code given at right-bottom of the figure. Scheme of inception modules are given at left-top of the figure.}
\label{fig:IRSNet}
\end{figure}

Network scheme can be seen in \ha{Figure} \ref{fig:IRSNet}. It has one encoder and two decoder streams to predict shading and reflectance simultaneously. Encoder has five inception module followed by $conv+pool$ layers to learn both shading and reflectance representations, and it has two decoders with 5 inception blocks followed by an upsampling layer. The output of the inception module in decoder is concatenated with respective encoder inception outputs. We used split-inception module in our network where the $n\times n$ convolutional filters are substituted by $n\times 1$ and $1\times n$ filters. This modification decreases overall learning parameters of the network that significantly results in a faster learning. 

Inspired by the architectures presented in DARN \cite{Lettry2018DARNAD} and IntrinsicNet \cite{Baslamisli18} we define a physical loss that constraints the solution to fulfil the intrinsic image model. We propose a unique global loss as a weighted sum of three terms, given by

\begin{equation}
\mathcal{L}_{Int}(I,\hat{R},\hat{S}) = 
\alpha_1 \mathcal{L}_{Ref}(R,\hat{R}) +
\alpha_2 \mathcal{L}_{Sha}(S,\hat{S}) +
\alpha_3 \mathcal{L}_{RS}(I,\hat{R}\cdot\hat{S})
\end{equation}

\noindent where the first two terms, $\mathcal{L}_{Ref}$ and $\mathcal{L}_{Sha}$, are two terms ensuring that Reflectance, $\hat{R}$, and Shading, $\hat{S}$, predictions fit the ground-truth data, respectively. The third term, $\mathcal{L}_{RS}$ forces the predictions to hold the intrinsic product model of \ha{Equation} \ref{eq:intrinsicdecomposition} and bound both outputs. All three terms are computing the mean square error (MSE) between the two inputs. It measures per pixel squared error between predicted images and the input ground-truth data, $R$, $S$ and $I$, for reflectance, shading and original image, respectively; $\alpha_i$ are the corresponding weights for each term. Our proposed network can be easily extended by stacking more decoder streams for each output and introducing more constrains in the loss function.

\section{Experiments and Results}\label{sec:experiments}

In this section we evaluate our IUI-Network architecture trained on our proposed dataset. We show that the proposed framework formed by a flexible architecture trained on a synthetic dataset with a wide range of variations helps to increase the quality of the results on generic intrinsic image estimation. To evaluate the approach, we compare our performance on existing datasets, either synthetic or real, and we compare to different methods. We show that with this preliminary dataset we achieve state of the art results.

In the following sections we perform 5 different experiments. In experiment 1, we show the results of our framework trained and tested on our dataset overpasses the performance of Retinex algorithm \cite{grosse09intrinsic}, which is a baseline method, still considered one of most effective methods for MIT dataset. In experiment 2 we provide the results on MIT dataset, which is only formed by real images. Results show that our fine-tuned network is able to get near state of art results. \ma{Similarly, in experiments 3 we show the effectiveness of our IUI network as it gives good results on Sintel dataset.}. In experiment 4 we provide a qualitative comparison on IIW dataset, results show that our network is able to capture texture details better than other methods. \ha{Finally, in experiment 5 we test the performance of the IUI architecture on Shapenet-Intrinsic dataset independently of our dataset} 

Before to start with the mentioned experiment we first give some implementation details and explain the error metrics, which is the standard used in previous works.

\subsection{Implementation details}

We implemented our network on \ha{Keras} \cite{chollet2015keras}. Weights were initialized using He Normal \cite{He2016DeepRL}. We used Adam optimizer \cite{AdamKingmaB14} with initial learning rate $0.0002$ which is decreased with factor of $0.1$ on reaching plateau. Since our network is all based on convolution layers, it enables to take any image size at the input. For our dataset the image size was decreased to $256\times256$ and for Sintel\cite{Butler:ECCV:2012:Sintel} we used $192\times448$. We used batch size of 8 for all experiments and used three dropout layers per decoding stream with $50\%$ dropout rate.  Our intrinsic loss computes mean square error between the real and the predicted shading and reflectance, and the product of these two with the original image. We initialized with equal weights for all $\alpha_i$ in our loss function. Network was trained from scratch with our dataset and fine tuned for others like MPI Sintel \cite{Butler:ECCV:2012:Sintel} and MIT \cite{grosse09intrinsic}. We also tested our network on IIW \cite{bell14intrinsic} images and provided results just for qualitative purpose.

\begin{figure}[h!]
\centering\includegraphics[width=13cm]{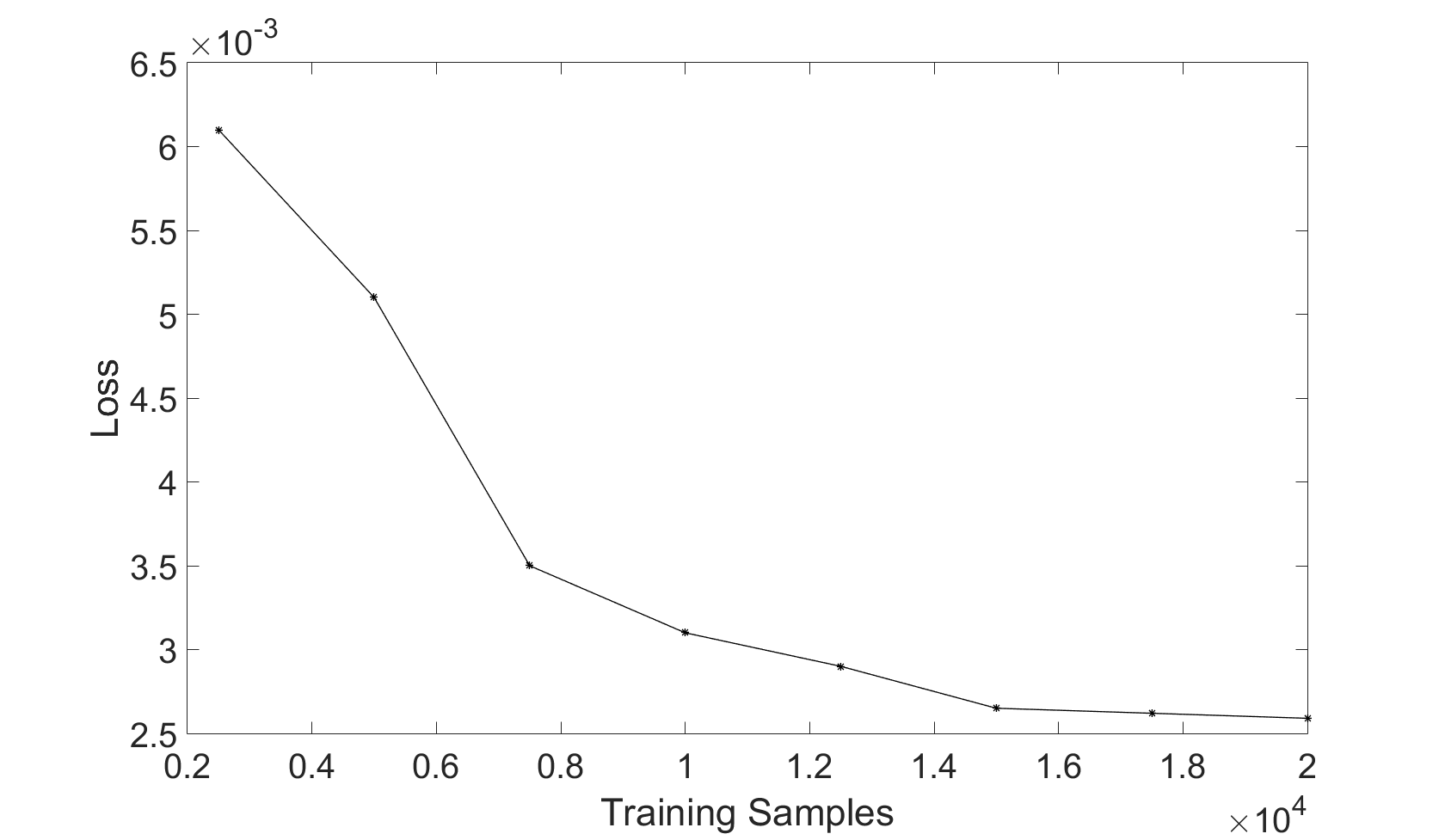}
\caption{Training vs. Loss for different training set sizes. Error stabilizes from 15,000 training samples.  }
\label{fig:training_vs_losst}
\end{figure}

We randomly split our dataset in $60\%$ of images used for training and $40\%$ used for testing. In \ha{Figure} \ref{fig:training_vs_losst} we show why we used this train/test split. We show the computed loss by different size of training sets, we can see that from $15000$ images we start to get a good intrinsic decomposition error, and increasing this number does not reduce the loss with a significant factor. We used $50$ epochs for computing the loss in this experiment. This split size could require to be changed if we would change the composition of the dataset by using a different set of parameters in the image generation stage. Thus, in all experiments based on our dataset, we trained the IUI-Network on 15,000 images and tested on 10,000 images. We compute the performance of our architecture on MPI Sintel \cite{Butler:ECCV:2012:Sintel} and MIT \cite{grosse09intrinsic}. Due to their small dimension or high redundancy, training a deep model directly on them is not feasible. To compare the feasibility of our approach with previous ones, we test on the existing datasets after specifically fine-tuning our trained IUI-Network.

\subsection{Error Metrics}
There are three most used metrics for Intrinsic image evaluation and comparison. Mean-squared error (MSE) and the local mean-squared error (LMSE) are more data-related metrics and structural dissimilarity index (DSSIM) is more a perceptual metric. In the formulation below,  $x$ denotes a ground-truth image and $\Bar{x}$ denotes predicted image. 

\textit{Mean-square error (MSE)} measures the mean square difference between estimated and ground-truth pixels in reflectance, shading or the product of these. Similar to \cite{grosse09intrinsic,chen2013simple,Narihira2015}, we also used scale-invariant metric to remove absolute brightness effect:

\begin{equation} 
 MSE(x,\Bar{x}) = \sum_{i=1}^{N}{\frac{||x_{i} - \Bar{\alpha} \Bar{x_{i}}||^{2} }{N}},
 \label{eq:MSE}
\end{equation}

\noindent $x$ can be reflectance $R$, shading $S$ or their product, $R\cdot S$, and $N$ is the total number of pixels in the evaluated estimation. The factor $\Bar{\alpha}=argmin_\alpha ||x_{i} - \alpha \Bar{x_{i}}||^{2}$ adjusts the absolute brightness between estimation and ground-truth to minimize the error and have scale invariance in metric.

\textit{Local mean-squared error (LMSE)} is used to compute scale-invariant MSE on overlapping windows of  $10 \%$ of the image size. If image is not square, the larger dimension is used to compute windows size.

\textit{Structural dissimilarity index (DSSIM)} is a dissimilarity version of the structural similarity index (SSIM). Differently, from the other metrics that measure absolute error between pixel values, SSIM is based on structural differences. It is motivated by human perception for structural difference.

\begin{equation} 
 SSIM(x, \Bar{x}) = \frac{ ( 2 \mu \Bar{\mu} + c_{1} )( 2 \sigma_{x\Bar{x}} + c_{2} ) }{ ( \mu^2 + \Bar{\mu}^2 + c_{1} ) (\sigma^2 + \hat{\sigma}^2 + c_{2})},
 \label{eq:ssim}
\end{equation}
\noindent where $\mu$, $\sigma^2$ and $\sigma_{x\hat{x}}$ are mean, variance and co-variance respectively; $c_{1}$ and $c_{2}$ are used to control zero approaching denominator. DSSIM is defined as

\begin{equation} 
 DSSIM(x,\Bar{x}) = \frac{1 - SSIM(x, \Bar{x})}{2}.
 \label{eq:dssim}
\end{equation}

\subsection{Experiment 1. Our dataset}

In this first experiment, we compared IUI-Network with Retinex\cite{grosse09intrinsic} for intrinsic decomposition. Major reason for choosing retinex on any other method is that it is easily available and it is the state of the art on MIT dataset. Results depicts that our network shows much improvement as compared to Retinex method. We computed the error separately on background walls, foreground objects and complete images. We show the results in \ha{Table} \ref{tab:ourdataset}.

We can see that our IUI-Network considerably improves the predictions of intrinsic decomposition for all parts of images dividing the error by a factor that goes from 5 to 50. In general the error improvement is better for shading than for reflectance, since the dataset present a strong bias on flat walls presenting large areas of homogeneous shading. This agrees with the fact that the error decreases more on backgrounds than foregrounds. The main reason for a better performance on the background is that foreground object has more diversified texture and shape which results in more shading and reflectance variation rather then background.

As mentioned, we get the best improvements on backgrounds for all metrics except for DSSIM in shading. Our intuition for better DSSIM for foreground rather than background is because of shadows. They are a structural part of background shading and our network is better in getting the structural shading of the objects than of cast shadows. To improve the recovery of the cast shadow structure in our future dataset we plan to diversify backgrounds using different textures for each wall and more diverse shapes than just flat walls. Introducing more randomly positioned multiple objects would reduce the bias that favour the background prediction and to use multiple color light source will make more challenging intrinsic image dataset for whole parts of image. 

In Figure \ref{fig:our dataset} we show qualitative results on our dataset, while our method is able to get correct intrinsic decomposition in most of images, it tends to have some errors in removing shadows from more plain background in reflectance images and keeping all shadow details in shading image, it also shows some blurring around room corners in shading images. First $3$ images shows more succeful cases while last $3$ shows some false intrinsic decomposition.

\begin{table}[!h]
    \centering
\begin{tabular}{|l||c|c|c|c|c|c|}
        \hline
        \multirow{2}{*}{\textbf{Method (where tested)}} & \multicolumn{3}{|c}{\textbf{Reflectance}} & \multicolumn{3}{|c|}{\textbf{Shading}} \\\cline{2-7}
        & MSE & LMSE & DSSIM & MSE & LMSE & DSSIM\\
         \hline \hline
         \textbf{Retinex (whole image)}\cite{grosse09intrinsic} & 0.0500 & 0.049 & 0.17 & 0.0400 & 0.0403 & 0.24 \\ \hline
         \textbf{\textit{IUI} (foreground object)} & 0.0046 & 0.0038 & 0.029 & 0.0023 & 0.0020 & \textbf{0.0178} \\ 
         & \textit{(10.9)} & \textit{(12.9)} & \textit{(5.9)} & \textit{(17.4)} & \textit{(20.2)} & \textit{(13.5)} \\
         \textbf{\textit{IUI} (background walls)} &\textbf{ 0.0016} & \textbf{0.0014} & \textbf{0.019} & \textbf{0.0010} & \textbf{0.0008} & 0.023 \\
          & \textit{(31.3)} & \textit{(35.0)} & \textit{(8.9)} & \textit{(40.0)} & \textit{(50.4)} & \textit{(10.4)} \\
         \textbf{\textit{IUI} (whole image)} & 0.0020 & 0.0019 & 0.020 & 0.0011 & 0.0009 & 0.022 \\
          & \textit{(25.0)} & \textit{(25.8)} & \textit{(8.5)} & \textit{(36.4)} & \textit{(44.8)} & \textit{(10.9)} \\
         \hline
    \end{tabular} 
    \caption{Errors for reflectance and shading predictions on our dataset. Comparison between our IUI architecture and Retinex algorithm. IUI decreases the error of Retinex by the factor given in brackets. Errors are separately reported on object, on foreground and the on whole image.}\label{tab:ourdataset}
\end{table}

\newcommand\sizephoto{0.19}
\begin{figure*}[h]
\begin{center}
\setlength\tabcolsep{0.5pt}
\begin{tabular}{ccccc}
\includegraphics[width=\sizephoto\textwidth]{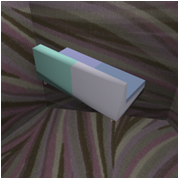} &
\includegraphics[width=\sizephoto\textwidth]{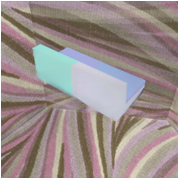} &
\includegraphics[width=\sizephoto\textwidth]{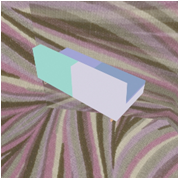} &
\includegraphics[width=\sizephoto\textwidth]{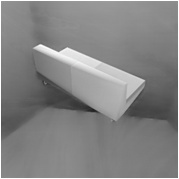} &
\includegraphics[width=\sizephoto\textwidth]{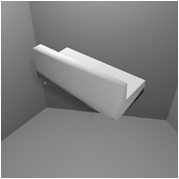} 
\\

\includegraphics[width=\sizephoto\textwidth]{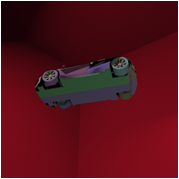} &
\includegraphics[width=\sizephoto\textwidth]{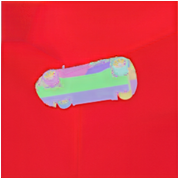} &
\includegraphics[width=\sizephoto\textwidth]{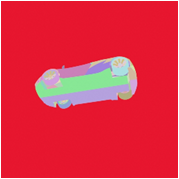} &
\includegraphics[width=\sizephoto\textwidth]{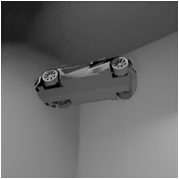} &
\includegraphics[width=\sizephoto\textwidth]{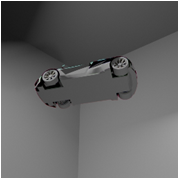} 
\\

\includegraphics[width=\sizephoto\textwidth]{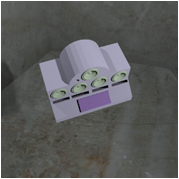} &
\includegraphics[width=\sizephoto\textwidth]{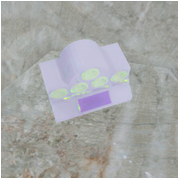} &
\includegraphics[width=\sizephoto\textwidth]{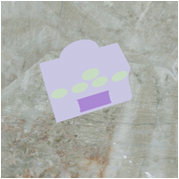} &
\includegraphics[width=\sizephoto\textwidth]{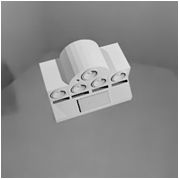} &
\includegraphics[width=\sizephoto\textwidth]{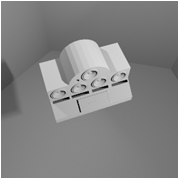} 
\\
\includegraphics[width=\sizephoto\textwidth]{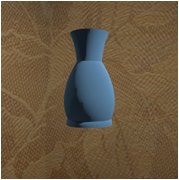} &
\includegraphics[width=\sizephoto\textwidth]{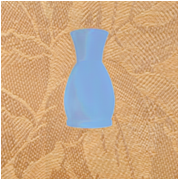} &
\includegraphics[width=\sizephoto\textwidth]{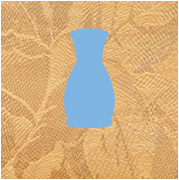} &
\includegraphics[width=\sizephoto\textwidth]{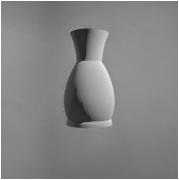} &
\includegraphics[width=\sizephoto\textwidth]{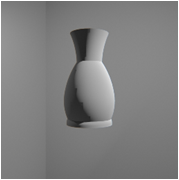} 
\\
\includegraphics[width=\sizephoto\textwidth]{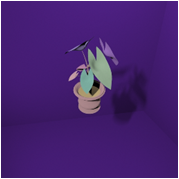} &
\includegraphics[width=\sizephoto\textwidth]{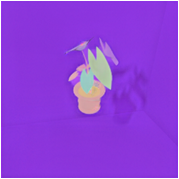} &
\includegraphics[width=\sizephoto\textwidth]{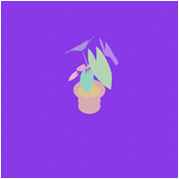} &
\includegraphics[width=\sizephoto\textwidth]{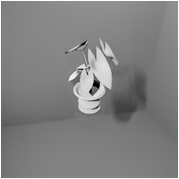} &
\includegraphics[width=\sizephoto\textwidth]{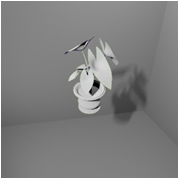} 
\\
\includegraphics[width=\sizephoto\textwidth]{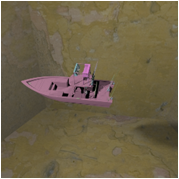} &
\includegraphics[width=\sizephoto\textwidth]{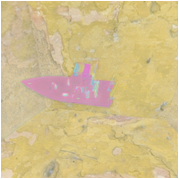} &
\includegraphics[width=\sizephoto\textwidth]{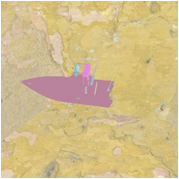} &
\includegraphics[width=\sizephoto\textwidth]{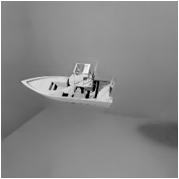} &
\includegraphics[width=\sizephoto\textwidth]{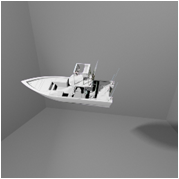} 
\\

(a) Input image &  (b) Ref. Predict. & (c) Ref. GT & (d) Shad. Predict. & (e) Shad. GT  \\


\end{tabular}
\end{center}
\caption{Some examples of our SID dataset. (a) Original Images. (b) and (d) are Reflectance and Shading estimation by our IUI network, respectivelly. (c) and (e) are GT Reflectance and Shading, repesctivelly.}
\label{fig:our dataset}
\end{figure*}

\subsection{Experiment 2. MIT dataset}

In this second experiment we evaluated IUI-Network on one of the most used dataset, MIT Intrinsic \cite{grosse09intrinsic}, and showed that it achieves nearly state of art results in all matrices. Similar to what all deep learning methods do, we fine-tuned our network from previous experiment to decompose the image in reflectance and shading, we used those weights as initialisation. Results on MIT dataset are given in \ha{Table} \ref{tab:MIT_Result}, the error is computed on the foreground mask to allow comparison with the rest of results reported in previous works.

Best performances are bold in the table. We achieve best errors for reflectance, and very close to InthinsicNet \cite{Baslamisli18} for Shading. These results confirm the generalization capability of our network and the effectiveness of training on our synthetic dataset that is able to generalize on  realistic images. In this case, our network was able to get better DSSIM results that  shows it is able to learn texture and edges information better than other methods. 

In \ha{Figure} \ref{fig:MIT Results} we show some results for visual comparison with other state of art methods. Our estimation of reflectance is giving  correct and sharp color information in dark shaded image areas. Regarding shading images we present a very good estimation for the turtle, and a worse on for the frog, where we can see some blurring effects and some shape details are lost.

\begin{table}[]
    \centering
\begin{tabular}{|l||c|c|c|c|c|c|}
        \hline
        \multirow{2}{*}{\textbf{Method}} & \multicolumn{3}{|c}{\textbf{Reflectance}} & \multicolumn{3}{|c|}{\textbf{Shading}} \\\cline{2-7}
        
        & MSE & LMSE & DSSIM & MSE & LMSE & DSSIM\\
         \hline \hline
         \textbf{Retinex}\cite{grosse09intrinsic} & \textbf{0.0032} & 0.0353 & 0.1825 & 0.0348 & 0.1027 & 0.3987 \\
         \textbf{SIRFS}\cite{Barron2015} & 0.0147 & 0.0416 & 0.1238 & 0.0083 & 0.0168 & 0.0985\\
         \textbf{Direct Intrinsics}\cite{Narihira2015} & 0.0277 & 0.0585 & 0.1526 & 0.0154 & 0.0295 & 0.1328 \\
         \textbf{ShapeNet}\cite{Shi2017LearningNO} & 0.0278 & 0.0503 & 0.1465 & 0.0126 & 0.0240 & 0.1200 \\
         \ha{\textbf{CGIntrinsics}\cite{li2018cgintrinsics}} & 0.167 & 0.0319 & 0.1287 & 0.0127 &0.0211 & 0.1376 \\
         \textbf{IntrinsicNet}\cite{Baslamisli18} & \textbf{0.0051} & 0.0295 & 0.0926 & \textbf{0.0029} &\textbf{ 0.0157} & \textbf{0.044}1 \\
         \textbf{RetiNet}\cite{Baslamisli18} & 0.0128 & 0.0652 & 0.0909 & 0.0107 & 0.0746 & 0.1054 \\
         \textbf{IUI} & \textbf{0.0046} & \textbf{0.0197} & \textbf{0.054} &0.0038 & 0.020 & 0.0557 \\
         \hline
    \end{tabular} 
    \caption{Estimation errors on MIT dataset reported in previous works by different methods and for our IUI architecture.}
    \label{tab:MIT_Result}
\end{table}
\newcommand\sizeMIT{1.7cm}

\begin{figure}[!ht]
\begin{center}
\setlength\tabcolsep{0.5pt}
\begin{tabular}{cccccccc}

\multirow{4}{*}[0.6cm]{\includegraphics[width=3.0cm]{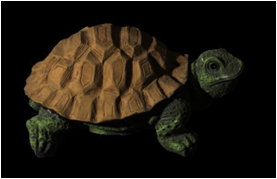}} &
\includegraphics[width=\sizeMIT]{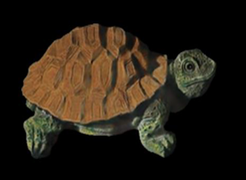} &
\includegraphics[width=\sizeMIT]{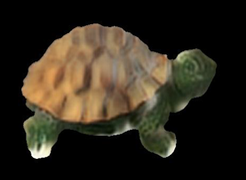} &
\includegraphics[width=\sizeMIT]{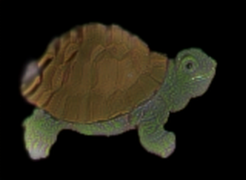} &
\includegraphics[width=\sizeMIT]{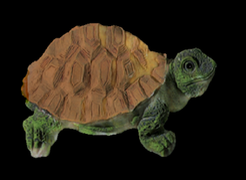} &
\includegraphics[width=\sizeMIT]{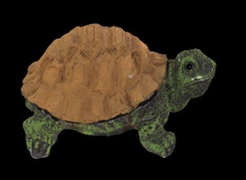} &
\raisebox{0.5cm}{\rotatebox[origin=c]{90}{\footnotesize{Reflectance}}}\\
~ & \includegraphics[width=\sizeMIT]{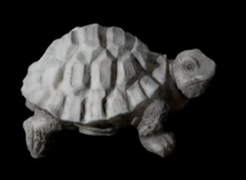} &
\includegraphics[width=\sizeMIT]{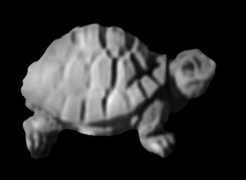} &
\includegraphics[width=\sizeMIT]{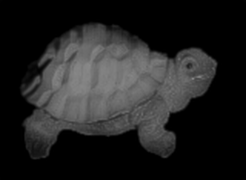} &
\includegraphics[width=\sizeMIT]{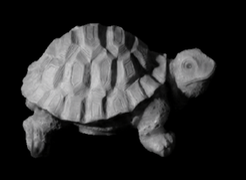} &
\includegraphics[width=\sizeMIT]{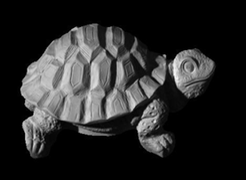}
&
\raisebox{0.5cm}{\rotatebox[origin=c]{90}{\footnotesize{Shading}}}
\\
\multirow{4}{*}[0.6cm]{\includegraphics[width=3.0cm]{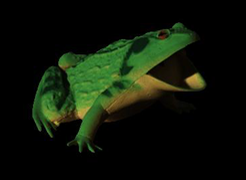}} &
\includegraphics[width=\sizeMIT]{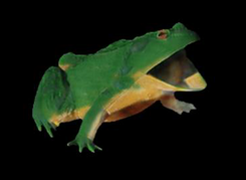} &
\includegraphics[width=\sizeMIT]{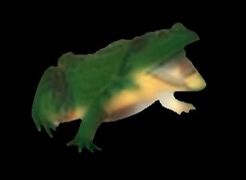} &
\includegraphics[width=\sizeMIT]{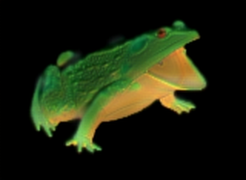} &
\includegraphics[width=\sizeMIT]{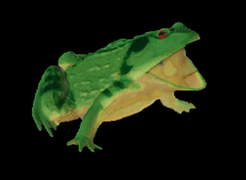} &
\includegraphics[width=\sizeMIT]{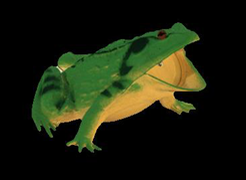} & \raisebox{0.5cm}{\rotatebox[origin=c]{90}{\footnotesize{Reflectance}}}\\ ~ &
\includegraphics[width=\sizeMIT]{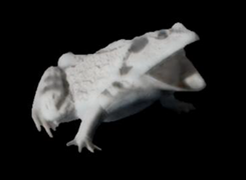} &
\includegraphics[width=\sizeMIT]{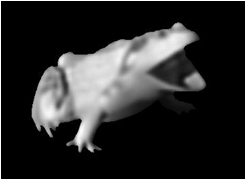} &
\includegraphics[width=\sizeMIT]{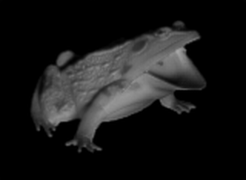} &
\includegraphics[width=\sizeMIT]{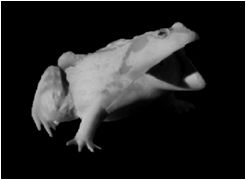} &
\includegraphics[width=\sizeMIT]{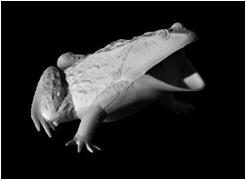} &
\raisebox{0.5cm}{\rotatebox[origin=c]{90}{\footnotesize{Shading}}}
\\
\footnotesize{Input} & 
\footnotesize{Shi\textit{ et al.}\cite{Shi2017LearningNO}} & \footnotesize{DI\cite{Narihira2015}} & \footnotesize{Intrinsic-Net\cite{Baslamisli18}} & \scriptsize{Ours} & \footnotesize{GT} & 
\\
\end{tabular}
\end{center}
\label{fig:MIT Results}
\caption{Qualitative Results on MIT intrinsic image dataset, compared to other methods, we achieved sharp and better colors and removed shading effects. Our method performed best in bringing reflectance details from dark part of image.}
\end{figure}

\subsection{Experiment 3. Sintel dataset}

Similarly to previous experiments, here we evaluate the performance of our architecture on the Sintel dataset \cite{Butler:ECCV:2012:Sintel}.  We followed the approach introduced by Narihira \textit{et al.} in \cite{Narihira2015} and provide the results for both image-split and scene-split tests. In image-split the whole $890$ images of the dataset are randomly assigned, such that $50\%$ are used for training and the rest for testing. In scene split, training ant testing data are decided on the basis of scenes, rather than images, which makes it more difficult in terms of capability to generalize.

In \ha{Table} \ref{tab:Sintel_Image_Result} we can see the results of our network fine tuned on the Image split test, denoted as \textit{IUI fine-tuned on CS}, altogether with previously reported results. In this case we can see that our method is giving an error very close to the current state of the art which is Fan \textit{et al.} \cite{Fan2018RevisitingDI}, we highlight those numbers closer by less than 0.002 to the best reported error. In the same table we report the results of two more experiments. We denote as \textit{IUI without fine-tuning} the results of testing our network without any training on this test set, just to see the performance of our network just trained on our SID, in this case we can see that our approach is better than most of previous methods before Fan \textit{et al.} method. Finally, we give the results of one additional experiment, we refer to it as \textit{IUI fine-tuned on GLS}, in this case we test our network fine tuned on a new ground-truth. Considering that Sintel dataset is based on synthetic images where shading is colored, we built a new version of this dataset with a grey-level shading that fully accomplishes the dot product model of intrinsic image decomposition with a single channel shading and RGB reflectance. We do this test to evaluate how our network is performing on this constraint, since it is trained under a loss function that forces the product model to be fulfilled ($\mathcal{L}_{RS}$ loss) like in SID dataset. We can show that in this case, our network is overcoming the state of the art in reflectance estimation.

In \ha{Table} \ref{tab:Sintel_Scene_Result} we can see the results of our network tested in the same manner as before but now on the Image split test. We confirm similar results. We are close to state of the art when fine-tuned on the this Sintel test and we overcome them in reflectance when we fine-tune on a grey-level shading dataset.

In \ha{Figures} \ref{fig: Sintel Image} and \ref{fig: sintel scene split} we show some examples of our results for image split and scene split respectively. We can see that Fan \textit{et al.} and our IUI are showing be best results from a qualitative point of view, getting a sharp and correct intrinsic decomposition. We can observe some interesting details in these images. For example, estimated reflectance in image split \ha{Figure} \ref{fig: Sintel Image} our network is giving better sky, clouds and background wall reflectance decomposition as compared to all other methods, and we get correct rust effects on cart and right brownish hairs for women. 
Regarding, scene split experiments, our network is getting better details in some parts of the image, for example in \ha{Figure} \ref{fig: sintel scene split} estimated reflectance in left column is correctly separating most shading effects on background walls, floor and cast shadows. We also get correct reflectance for animal left and right eyes. Similarly, in right column, our estimation is getting best recovery of color details in teeth, mouth and arm colors. 

However, our method does not perform accurately in recovering shading images and it shows overall false color for shading detection. We believe the main reason behind this issue is that our method is trained on our current SID version that only contains white light sources, while Sintel scenes present some unnatural lights with bias towards bluish or brownish colors. This open a further research line in our framework which is introducing more color light sources in SID. Finally, we also want to point out, that on the shading of left column image in \ha{Figure} \ref{fig: sintel scene split}, our method recovers some specular effects on animal face which, less artifacts on floor area and a smoother background, all closer to the ground-truth shading. 

\begin{table}[ht!]
    \centering
\begin{tabular}{|l||c|c|c|c|c|c|}
        \hline
        \multirow{2}{*}{\textbf{Method}} & \multicolumn{3}{|c}{\textbf{Reflectance}} & \multicolumn{3}{|c|}{\textbf{Shading}} \\\cline{2-7}
        
        & MSE & LMSE & DSSIM & MSE & LMSE & DSSIM\\
         \hline \hline
         \textbf{Retinex}\cite{grosse09intrinsic} & 0.0606 & \textbf{0.00366} & 0.227 & 0.0727 & 0.0419 & 0.24 \\
         \textbf{Lee \textit{et al.}}\cite{Lee2012} & 0.0463 & 0.02224 & 0.199 & 0.0507 & 0.0192 & 0.177 \\
         \textbf{SIRFS}\cite{Barron2015} & 0.042 & 0.0298 & 0.21 & 0.0436 & 0.0264 & 0.206 \\
         \textbf{Chen and Koltun}\cite{chen2013simple} & 0.0307 & 0.0185 & 0.196 & 0.0277 & 0.019 & 0.165 \\
         \textbf{Direct Intrinsics}\cite{Narihira2015} & 0.01 & 0.0083 & \textbf{0.02014} & 0.0092 & 0.0085 & 0.1505 \\
         \textbf{Fan \textit{et al.}}\cite{Fan2018RevisitingDI} & \textbf{0.0069} & \textbf{0.0044} & \textbf{0.1194} & \textbf{0.0059} &\textbf{ 0.0043} & \textbf{0.0822} \\
         \textbf{IUI fine-tuned on CS} & \textbf{0.0072} & \textbf{0.0054} & 0.1374 & 0.0068 & \textbf{0.0059} & 0.1247 \\ \hline
          \textbf{IUI without fine-tuning} & 0.023 & 0.015 & 0.21 & 0.035 & 0.022 & 0.255 \\
         \textbf{IUI fine-tuned on GLS} &\textbf{ 0.0062} & \textbf{0.0047} & 0.1297 & \textbf{0.0057} & \textbf{0.0048} & 0.1183 \\ 
         \hline
    \end{tabular} 
    \caption{Results on Sintel Image Split dataset. Best errors are highlighted in bold.}
    \label{tab:Sintel_Image_Result}
\end{table}

\begin{table}[!ht]
    \centering
\begin{tabular}{|l||c|c|c|c|c|c|}
        \hline
        \multirow{2}{*}{\textbf{Method}} & \multicolumn{3}{|c}{\textbf{Reflectance}} & \multicolumn{3}{|c|}{\textbf{Shading}} \\\cline{2-7}
        
        & MSE & LMSE & DSSIM & MSE & LMSE & DSSIM\\
         \hline \hline
         \textbf{Direct Intrinsics}\cite{Narihira2015} & 0.0238 & 0.0155 & 0.226 & 0.0205 & 0.0172 & 0.1816 \\
         \textbf{Fan \textit{et al.}}\cite{Fan2018RevisitingDI} & \textbf{0.0189} & \textbf{0.0122} & \textbf{0.1645} & \textbf{0.0171} & \textbf{0.0117} & \textbf{0.1450 }\\
        \textbf{IUI fine tuned on CS} & 0.0213 & 0.0140 & 0.1787 & 0.0253 & 0.01721 & 0.1874 \\ \hline
        \textbf{IUI without fine-tuning} & 0.023 & 0.0154 & 0.20 & 0.034 & 0.023 & 0.24 \\ 
         \textbf{IUI fine tuned on GLS} & \textbf{0.01733} & \textbf{0.0110} & \textbf{0.16189} & 0.0201 & 0.013182 & 0.1618 \\ 
         \hline
    \end{tabular} 
    \caption{Result on Sintel Scene Split dataset. Best errors are highlighted in bold.}
    \label{tab:Sintel_Scene_Result}
\end{table}
\newcommand\sizesintel{0.24}
\begin{figure}[h]
\begin{center}
\setlength\tabcolsep{0.5pt}
\begin{tabular}{ccccc}
\raisebox{1.3cm}{\rotatebox[origin=c]{90}{\footnotesize{Input}}}& \multicolumn{2}{c}{\includegraphics[width=0.44\textwidth]{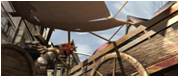}} &
\multicolumn{2}{c}{\includegraphics[width=0.44\textwidth]{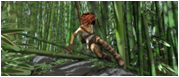}} \\
\raisebox{0.7cm}{\rotatebox[origin=c]{90}{\footnotesize{SIRFS\cite{Barron2015}}}}
 &
\includegraphics[width=\sizesintel\textwidth]{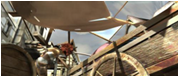} &
\includegraphics[width=\sizesintel\textwidth]{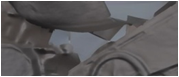} &
\includegraphics[width=\sizesintel\textwidth]{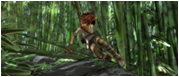} &
\includegraphics[width=\sizesintel\textwidth]{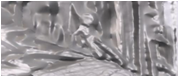} \\
\raisebox{0.65cm}{\rotatebox[origin=c]{90}{\footnotesize{Chen \textit{et al.}\cite{chen2013simple}}}}
 &
\includegraphics[width=\sizesintel\textwidth]{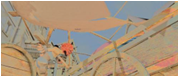} &
\includegraphics[width=\sizesintel\textwidth]{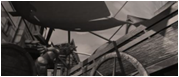} &
\includegraphics[width=\sizesintel\textwidth]{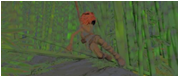} &
\includegraphics[width=\sizesintel\textwidth]{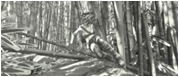} \\
\raisebox{0.6cm}{\rotatebox[origin=c]{90}{\footnotesize{DI \cite{Narihira2015}}}}
 &
\includegraphics[width=\sizesintel\textwidth]{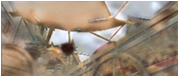} &
\includegraphics[width=\sizesintel\textwidth]{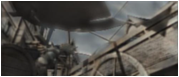} &
\includegraphics[width=\sizesintel\textwidth]{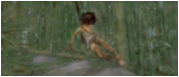} &
\includegraphics[width=\sizesintel\textwidth]{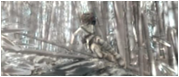} \\
\raisebox{0.7cm}{\rotatebox[origin=c]{90}{\footnotesize{Fan \textit{et al.}\cite{Fan2018RevisitingDI}}}}
&
\includegraphics[width=\sizesintel\textwidth]{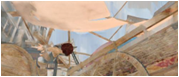} &
\includegraphics[width=\sizesintel\textwidth]{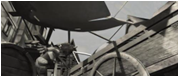} &
\includegraphics[width=\sizesintel\textwidth]{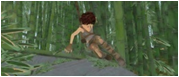} &
\includegraphics[width=\sizesintel\textwidth]{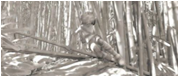} \\
\raisebox{0.7cm}{\rotatebox[origin=c]{90}{\footnotesize{IUI}}} &
\includegraphics[width=\sizesintel\textwidth]{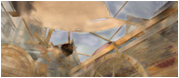} &
\includegraphics[width=\sizesintel\textwidth]{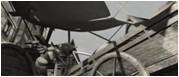} &
\includegraphics[width=\sizesintel\textwidth]{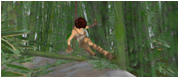} &
\includegraphics[width=\sizesintel\textwidth]{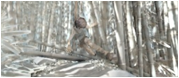} \\
\raisebox{0.7cm}{\rotatebox[origin=c]{90}{\footnotesize{GT}}}  &
\includegraphics[width=\sizesintel\textwidth]{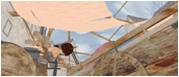} &
\includegraphics[width=\sizesintel\textwidth]{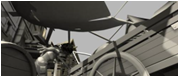} &
\includegraphics[width=\sizesintel\textwidth]{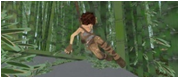} &
\includegraphics[width=\sizesintel\textwidth]{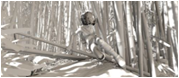} \\
& Reflectance & Shading & Reflectance & Shading \\

\end{tabular}
\end{center}
\caption{Visual comparison on MPI-Sintelt dataset using image split.}
\label{fig: Sintel Image}
\end{figure}

\begin{figure}[!ht]
\begin{center}
\setlength\tabcolsep{0.5pt}
\begin{tabular}{lcccc}
\raisebox{1.2cm}{\rotatebox[origin=c]{90}{\footnotesize{Input}}} & \multicolumn{2}{c}{\includegraphics[width=0.44\textwidth]{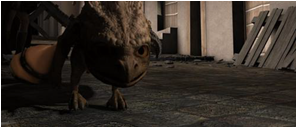}} &
\multicolumn{2}{c}{\includegraphics[width=0.44\textwidth]{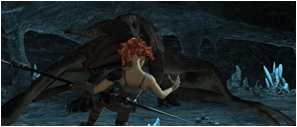}} \\
\raisebox{0.8cm}{\rotatebox[origin=c]{90}{\footnotesize{DI \cite{Narihira2015}}}} &
\includegraphics[width=\sizesintel\textwidth]{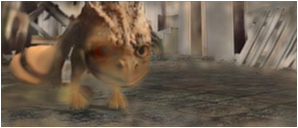} &
\includegraphics[width=\sizesintel\textwidth]{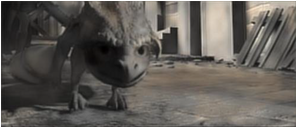} &
\includegraphics[width=\sizesintel\textwidth]{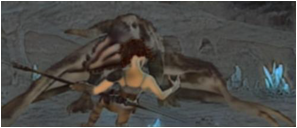} &
\includegraphics[width=\sizesintel\textwidth]{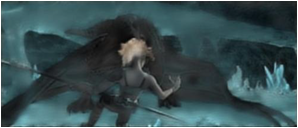} \\
\raisebox{0.7cm}{\rotatebox[origin=c]{90}{\footnotesize{Fan \textit{et al.}\cite{Fan2018RevisitingDI}}}}&
\includegraphics[width=\sizesintel\textwidth]{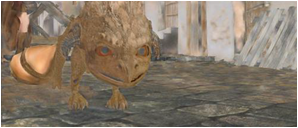} &
\includegraphics[width=\sizesintel\textwidth]{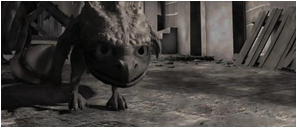} &
\includegraphics[width=\sizesintel\textwidth]{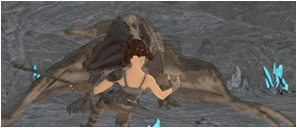} &
\includegraphics[width=\sizesintel\textwidth]{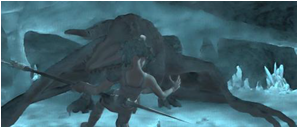} \\
\raisebox{0.6cm}{\rotatebox[origin=c]{90}{\footnotesize{IUI}}} &
\includegraphics[width=\sizesintel\textwidth]{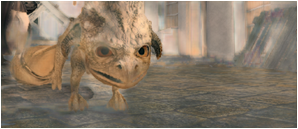} &
\includegraphics[width=\sizesintel\textwidth]{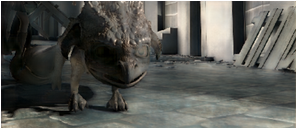} &
\includegraphics[width=\sizesintel\textwidth]{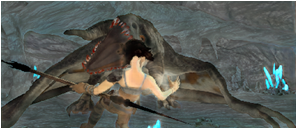} &
\includegraphics[width=\sizesintel\textwidth]{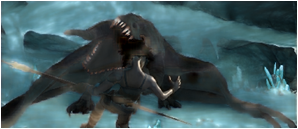} \\
\raisebox{0.6cm}{\rotatebox[origin=c]{90}{\footnotesize{GT}}}  &
\includegraphics[width=\sizesintel\textwidth]{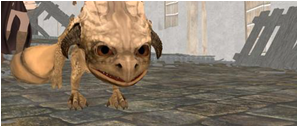} &
\includegraphics[width=\sizesintel\textwidth]{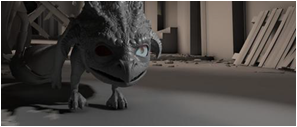} &
\includegraphics[width=\sizesintel\textwidth]{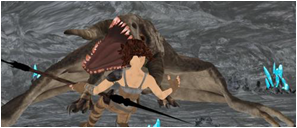} &
\includegraphics[width=\sizesintel\textwidth]{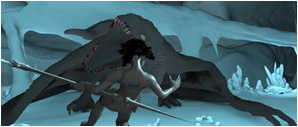} \\
& Reflectance & Shading & Reflectance & Shading \\
\end{tabular}
\end{center}
\caption{Visual comparison on MPI-Sintel using Scene split.}
\label{fig: sintel scene split}
\end{figure}

\subsection{Experiment 4. IIW dataset}

In this last section we provide a qualitative experiment on real images. Although, this is a very preliminar experiment, since we did not focused on training our approach on this dataset, we want to add here two interesting observations: (a) the main strength of our architecture is the ability of recovering texture information in the reflectance component; and (b) the proposed architecture is based on a very simple and versatile approach that presents a very low computational cost compared to the rest of state of art methods.

\begin{figure*}[!ht]
\includegraphics[width=0.9\textwidth]{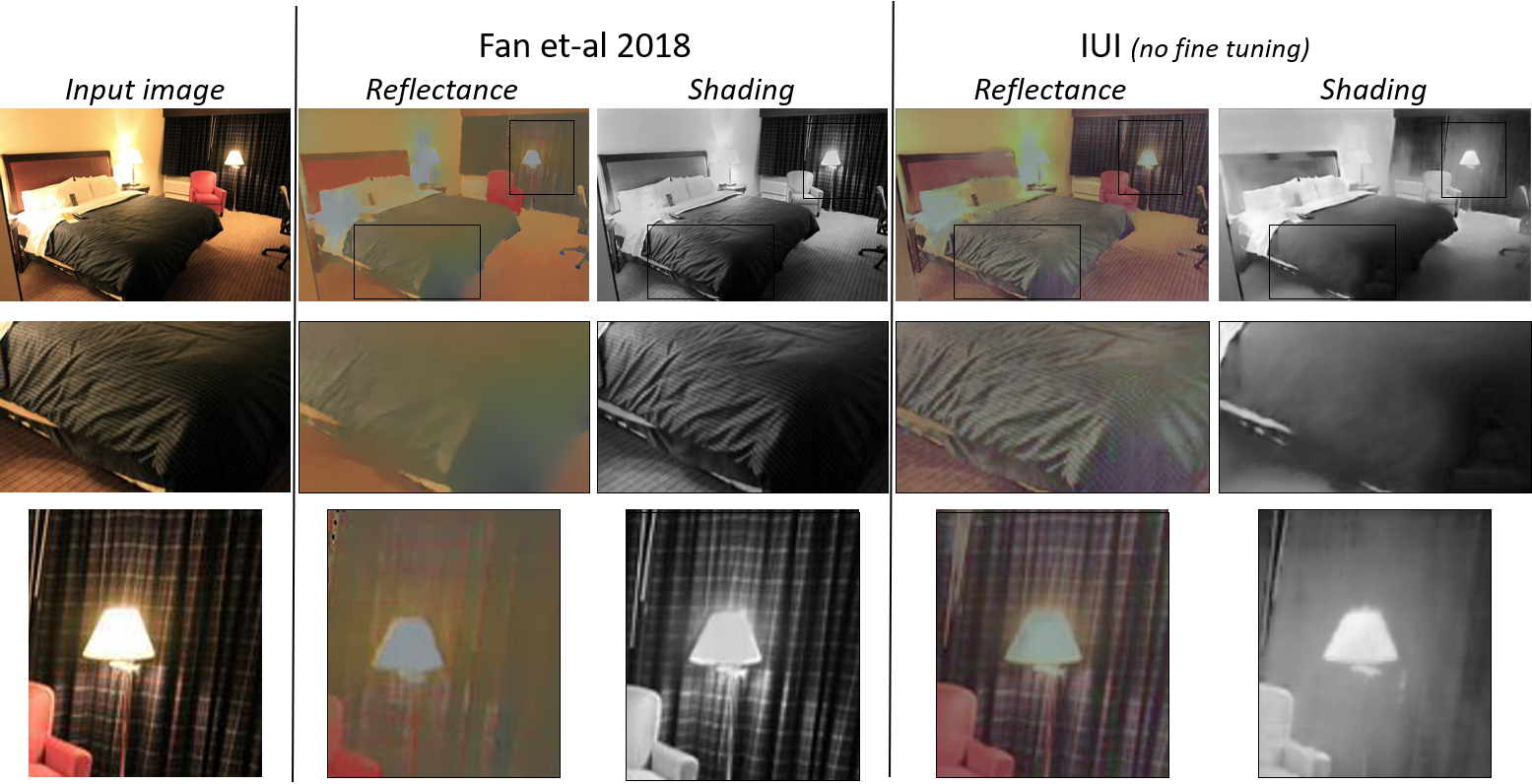}
\caption{Qualitative results on IIW}
\label{fig:IIW}
\end{figure*}

In \ha{Figure} \ref{fig:IIW} we show an example to compare between one of most recent and efficient methods with real images, which is Fan \textit{et al.} \cite{Fan2018RevisitingDI} trained on this dataset; and our IUI architecture without fine tuning on the dataset. The predicted images shown in the figure are by direct application of IUI trained on our SID dataset.
In this example we can see that at the level of the object textures, our approach is providing a better decomposition, reflectance prediction is capturing all the texture details of blanket, carpet and curtain, while we fail in getting unsharp edges for these objects. However, the decomposition provided by Fan \textit{et al.} results is getting very good sharp versions of Shading while including a lot of details that should be in the reflectance image. Therefore, some more work is required in setting more adapted versions of our dataset to overcome these problems, but, we think our framework is providing an excellent tool to overcome the difficulties of a good intrinsic decomposition at all levels.  

In \ha{Table} \ref{tab:IIW_dataset_Result} we show two quantitative aspects we can conclude from experiments on IIW dataset. We have to remind that this dataset is based on human judgments about shading and it is providing poor information about true reflectance and its spatial coherence. Although our network was not trained for IIW dataset, and we have proved our approach is better in getting reflectance estimation,  still we get $22.50 \%$ accuracy between our shading predictions and those provided by human judgments. In the same table, we also list computation times for several methods and we show that our inception-based network, and thanks to its small number of parameters, is giving the second best performance. This table numbers have been reproduced from \cite{Fan2018RevisitingDI}

\begin{table}[!ht]
    \centering
\begin{tabular}{|c||c|c|}
        \hline
         Method & WHDR(mean) & runtime(sec)\\
         \hline \hline
         \textbf{Shen \textit{et al.}}\cite{ShenIntrinsic2011} & 36.90 & 297\\
         \textbf{Retinex(color)}\cite{grosse09intrinsic} & 26.89 & 198.5\\
         \textbf{Retinex(gray)}\cite{grosse09intrinsic} & 26.84 & 225.3\\
         \textbf{Graces \textit{et al.}}\cite{Garces2012IntrinsicClustring} & 25.46 & 5.1\\
         \textbf{Zhao \textit{et al.}}\cite{Zhao:2012:Closed-Form} & 23.20 & 34.7\\
         \textbf{IUI} \textit{(without fine-tuning)} & 22.50 & \textbf{0.02}\\
         \textbf{\textit{L1} flattening}\cite{BIL1Image2015} & 20.94 & 310.94\\
         \textbf{Bell \textit{et al.}}\cite{bell14intrinsic} & 20.64 & 214\\
         \textbf{Zhou \textit{et al.}}\cite{Zhou2015LearningDR} & 19.95 &  300\\
         \textbf{Nestmeyer \textit{et al.} (CNN)}\cite{Nestmeyer2017ReflectanceAF} & 19.49 & \textbf{0.006}\\
         \textbf{Nestmeyer \textit{et al.}}\cite{Nestmeyer2017ReflectanceAF} & 17.69 & 300.086 \\
         \textbf{Bi \textit{et al.}}\cite{BIL1Image2015} & 17.67 & 300\\
         \textbf{Fan \textit{et al.}}\cite{Fan2018RevisitingDI} & \textbf{14.45} & 0.1\\
         \hline
    \end{tabular} 
    \caption{Result on IIW dataset}
    \label{tab:IIW_dataset_Result}
\end{table}

\ha{\subsection{Experiment 5. \ma{Evaluating IUI architecture}}} 

\ma{The goal of this experiment is to test the performance of the IUI architecture independently of our dataset. We trained our IUI-Network from scratch on the ShapeNet-Intrinsic dataset introduced by Baslamisli \textit{et al.}. We followed the same training/testing split introduced in their work and we used the same metrics to compare performance. Results reported in Table \ref{tab:Shapenet-Intrinsic} show that our architecture and the dataset help to improve  results. In the second experiment, we trained and tested our IUI-Network on the ShapeNet-Intrinsic dataset from scratch.Results show that our network outperforms state of art methods on this specific dataset. Best performances are bold faced in the table.} 

\ma{Additionally, we performed another experiment that we called \textit{IUI without fine-tuning}. We trained our network on our own SID dataset and tested it on shapeNet-Intrinsic dataset. Our network without fine-tuning performed better than Direct-Intrinsics\cite{Narihira2015} fine-tuned on this dataset. This experiment, likewise those reported in previous sections, shows a good generalization capability of our IUI network combined with our SID dataset.}

\begin{table}[h!]
    \centering
    \resizebox{\columnwidth}{!}{
\color{black}\begin{tabular}{|l||c|c|c|c|c|c|}
        \hline
        \multirow{2}{*}{\textbf{Method}} & \multicolumn{3}{|c}{\textbf{Reflectance}} & \multicolumn{3}{|c|}{\textbf{Shading}} \\\cline{2-7}
        
        & MSE & LMSE & DSSIM & MSE & LMSE & DSSIM\\
         \hline \hline
         \textbf{Direct Intrinsic} \cite{Narihira2015} & 0.1487 & 0.6868 & 0.0475 & 0.0505 & 0.3386 & 0.0361 \\
         \textbf{ShapeNet} \cite{Shi2017LearningNO} & 0.0023 & 0.0349 & 0.0186 & 0.0037 & 0.0608 & 0.0171 \\
         \textbf{IntrinsicNet} \cite{Baslamisli18} & 0.0005 & 0.0072 & 0.0909 & 0.0007 & 0.0505 & 0.0084 \\ 
         \textbf{RetiNet} \cite{Baslamisli18} & 0.0003 & 0.0205 & 0.0052 & 0.0004 & 0.0253 & 0.0064 \\
         \textbf{IUI \textit{trained on Shapenet-\ma{Intrinsic}}} & \textbf{0.0002} & \textbf{0.0193} & \textbf{0.0032} & \textbf{0.0003} & \textbf{0.0229} & \textbf{0.0047} \\ \hline
        \textbf{IUI \textit{without fine-tuning}} & 0.0073 & 0.1926 & 0.0396 & 0.0479 & 0.2324 & 0.0291 \\ 
         \hline
    \end{tabular}} 
    \caption{\ha{Estimation errors of different architectures trained on Shapenet-Intrinsic dataset. In bottom row the errors of IUI architecture trained on our dataset and tested on Shapenet-Intrinsic.}}
    \label{tab:Shapenet-Intrinsic}
\end{table}

\section{Conclusion}\label{sec:conclusions}

In this work we propose a versatile framework to define and train a convolutional network able to perform a intrinsic decomposition through training on a dataset with a large variety of light effects and color reflectances. The approach presented and evaluated here is a first version, where we have just worked with single white light sources, single background, and a limited number of room shapes, all of them based on flat surfaces. A wide range of variations can be introduced to improve the diversity of the scenes to be trained on. 

In parallel our proposed CNN architecture has been defined in a simplistic way to reduce its number of parameters and enough flexible to be adapted to multiple type of visual tasks related to light effect estimation. Apart from intrinsic decomposition it can be easily extended to color constancy or cast shadow removal, we already have preliminary results on these fields. 

The results obtained by all the experiments we report in this paper, make us to be optimistic about the capabilities of the presented approach to train networks devoted to solve \ma{taks related to the estimation of} light effects. In all the reported experiments we show a performance close to the state of the art of the problem of \ma{intrinsic decomposition in shading and reflectance.}

\section*{Acknowledgments} 
This work has been supported project TIN2014-61068-R and FPI Predoctoral Grant (BES-2015-073722), project RTI2018-095645-B-C21 of Spanish Ministry of Economy and Competitiveness and the CERCA Programme / Generalitat de Catalunya.

\bibliography{sample}

\end{document}